\documentclass[runningheads,a4paper]{llncs}
\usepackage{graphicx}
\usepackage{textcomp}
\usepackage[misc,geometry]{ifsym} 
\usepackage{url}
\urldef{\mailsa}\path|{alexander.hagg, dominik.wilde, alexander.asteroth}@h-brs.de|    
\urldef{\mailsb}\path|t.h.w.baeck@liacs.leidenuniv.nl| 

\usepackage{amsmath}
\usepackage{mathtools}
\usepackage{amsfonts}
\usepackage{placeins}
\usepackage{algorithm,algorithmicx,algpseudocode}
\usepackage{threeparttable}
\usepackage{color}

\begin{document}
\title{Designing Air Flow with Surrogate-assisted Phenotypic Niching}
\authorrunning{A. Hagg et al.}
\author{Alexander Hagg\inst{1,3}\textsuperscript{(\Letter)}\orcidID{0000-0002-8668-1796} \and
	Dominik Wilde\inst{2,1}\orcidID{0000-0003-3263-7287} \and
	Alexander Asteroth\inst{1}\orcidID{0000-0003-1133-9424} \and
	Thomas B\"ack\inst{3}\orcidID{0000-0001-6768-1478}}

\institute{Bonn-Rhein-Sieg University of Applied Sciences, Sankt Augustin, Germany\\\mailsa \and
Chair of Fluid Mechanics, University of Siegen, Siegen, Germany\\ \and
	Leiden Institute of Advanced Computer Science, Leiden University, Leiden, The Netherlands\\\mailsb}
\maketitle              
\begin{abstract}
In complex, expensive optimization domains we often narrowly focus on finding high performing solutions, instead of expanding our understanding of the domain itself. But what if we could quickly understand the complex behaviors that can emerge in said domains instead? We introduce surrogate-assisted phenotypic niching, a quality diversity algorithm which allows to discover a large, diverse set of behaviors by using computationally expensive phenotypic features. In this work we discover the types of air flow in a 2D fluid dynamics optimization problem. A fast GPU-based fluid dynamics solver is used in conjunction with surrogate models to accurately predict fluid characteristics from the shapes that produce the air flow. We show that these features can be modeled in a data-driven way while sampling to improve performance, rather than explicitly sampling to improve feature models. Our method can reduce the need to run an infeasibly large set of simulations while still being able to design a large diversity of air flows and the shapes that cause them. Discovering diversity of behaviors helps engineers to better understand expensive domains and their solutions. 

\keywords{evolutionary computation \and quality diversity \and phenotypic niching \and computational fluid dynamics \and surrogate models \and Bayesian optimization.}
\end{abstract}

\section{Introduction}

We design objects with the expectation that they will exhibit a certain behavior. In fluid dynamics optimization, we want an airplane wing to experience low drag forces, but also have a particular lift profile, depending on angle of attack and air speed. We want to understand how the design of our public transportation hubs, dealing with large influxes of travelers, can cause congestion at maximal flow rates. We want our buildings to cause as little wind nuisance as possible and understand how their shape and the wind turbulence they cause are linked. In all these cases, it is not easy to design without prior experience and we often require long iterative design processes or trial-and-error methods. 

What if we could quickly understand the possible types of behavior in expensive engineering problems and get an early intuition about how shape and behavior are related? In this work, we try to answer these questions, and in particular, whether we can discover different high performing behaviors of shapes, designing air flow simultaneously to the shapes that causes it. An overview of related work is given in Section~\ref{sec:relatedwork}, where we explain quality diversity (QD) algorithms and the use of surrogate assistance. In Section~\ref{sec:method} we introduce a new QD algorithm that performs surrogate-assisted phenotypic niching. Two problem domains are used (Section~\ref{sec:domains}): one inexpensive domain that optimizes the symmetry of polygons, allowing us to perform an in depth evaluation of various QD methods, and an expensive air flow domain (Section~\ref{sec:evaluation}).

\section{Quality Diversity}
\label{sec:relatedwork}

QD algorithms combine performance based search with ``blind'' novelty search, which searches for novel solutions without taking into account performance~\cite{Lehman2011a}. QD finds a diverse set of high performing optimizers~\cite{Cully2015,Lehman2011} by only allowing solutions to compete in local niches. Niches are based on features that describe phenotypic aspects, like shape, structure or behavior. It keeps track of an archive of niches and solutions are added if their phenotype fills an empty niche or their quality is higher than that of the solution that was previously placed inside. 

QD became applicable to expensive optimization problems after the introduction of surrogate-assisted illumination (SAIL)~\cite{Gaier2017}. In this Bayesian interpretation of QD, a multi-dimensional archive of phenotypic elites (MAP-Elites)~\cite{Cully2015} is created based on \textit{upper confidence bound} (UCB)~\cite{Auer2003} sampling, which takes an optimistic view at surrogate-assisted optimization. A Gaussian Process (GP) regression~\cite{rasmussen1997} model predicts the performance of new solutions based on the distance to previous examples, which is modeled using a covariance function. A commonly used covariance function is the squared exponential, which has two hyperparameters: the length scale (or sphere of influence) and the signal variance, which are found by minimizing the negative loglikelihood of the process. For any location, the GP model predicts a mean value $\mu$ and confidence intervals $\sigma$ of the prediction. $\sigma$ is added to $\mu$ with the idea that a location where the model has low confidence also has the promise of holding a better performing solution: $UCB(x) = \mu(x) + \kappa \cdot \sigma(x)$. The parameter $\kappa$ allows us to tune the UCB function between exploitation ($\kappa = 0$) and exploration ($\kappa >> 0$).

In SAIL, after MAP-Elites fills the acquisition map which contains ``optimistic'' solution candidates, a random selection of those candidates is analyzed in the expensive evaluation function to form additional training samples for the GP model. This loop continues until the evaluation budget is exhausted. Then $\kappa$ is set to 0 in a final MAP-Elites run to create a feature map that now contains a diverse set of solutions that is predicted to be high-performing. SAIL needs a budget orders of magnitudes smaller than MAP-Elites because it can exploit the surrogate model without ``wasting'' samples. SAIL, however, is constrained to features that are cheap to calculate, like shape features~\cite{Gaier2017} that can be determined without running the expensive evaluation. 

With SAIL it became possible to use performance functions of expensive optimization domains. But the strength of QD, to perform niching based on behavior, cannot be applied when determining those behaviors is expensive. In this work we evaluate whether we can include surrogate models for such features.

\section{Surrogate-assisted Phenotypic Niching}
\label{sec:method}

\begin{figure}[tb]
	\centering
	\includegraphics{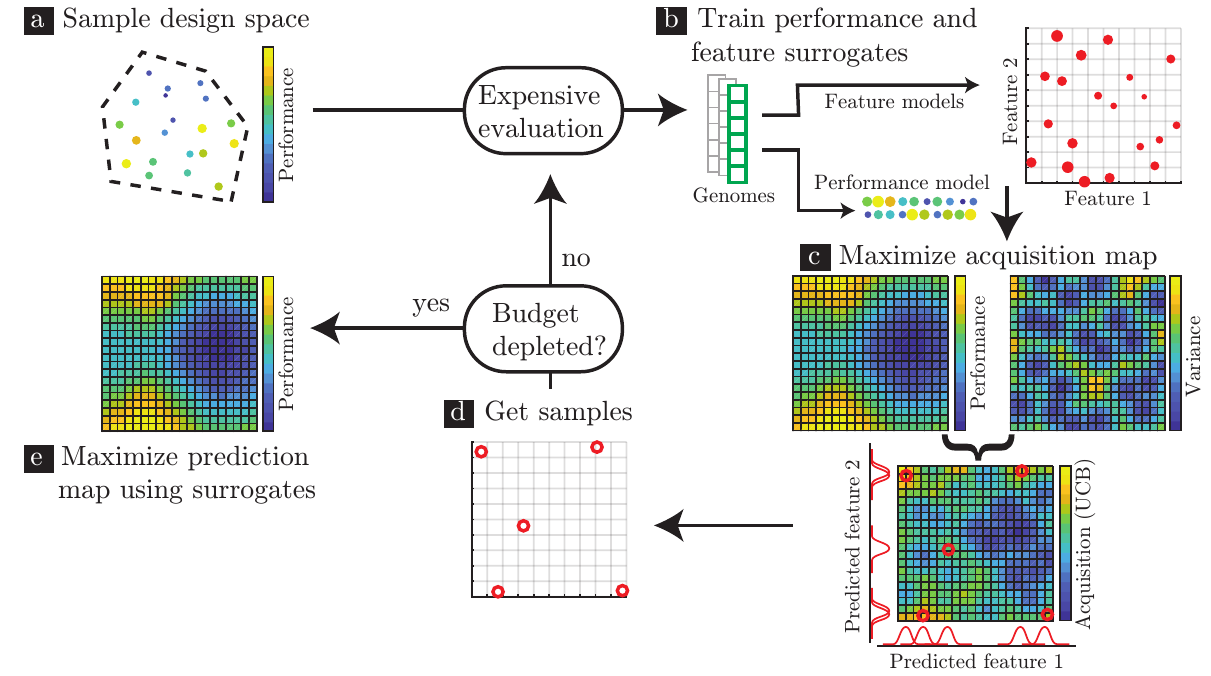}
	\caption{\textbf{Surrogate-assisted Phenotypic Niching.} An initial sample set is evaluated (a), then models are trained to predict performance and feature coordinates (b), MAP-Elites is used to produce an acquisition map, balancing exploitation and exploration with the UCB of the performance model. Feature models predict the niche of new individuals (c). New samples are selected from the acquisition map (d). After the evaluation budget is depleted, the surrogate models are used to generate the prediction map with MAP-Elites, ignoring model confidence (e).}
	\label{img:SPHEN}
\end{figure}

To be able to handle expensive features, we introduce surrogate-assisted phenotypic niching (SPHEN) (Fig.~\ref{img:SPHEN} and Alg.~\ref{alg:SPHEN}). By building on the insight that replacing the performance function with a surrogate model decreases the necessary evaluation budget, we replace the exact feature descriptors as well. 

The initial training sample set, used to form the first seeds of the acquisition map, is produced by a space-filling, quasi-random low-discrepancy Sobol~\cite{sobol1967distribution} sequence in the design space (Fig.~\ref{img:SPHEN}a). Due to the lack of prior knowledge in black-box optimization, using space-filling sequences has become a standard method to ensure a good coverage of the search domain. The initial set is evaluated, for example in a computational fluid dynamics simulator. Performance and phenotypic features of those samples are derived from the results, or, in the case of simpler non-behavioral features, from the solutions' expression or shape themselves. The key issue here is to check the range of the initial set's features. Since we do not know what part of the phenotypic feature space will be discovered in the process, the initial set's feature coordinates only give us a hint of the reachable feature space. Just because we used a space-filling sampling technique in the design space, does not mean the samples are space-filling in feature space. 

After collecting performance and feature values, the surrogate models are trained (Fig.~\ref{img:SPHEN}b). We use GP models, which limit the number of samples to around 1,000, as the training and prediction becomes quite expensive. A squared exponential covariance function is used and the (constant) mean function is set to the training samples' mean value. The covariance function's hyperparameters, length scale and signal variance, are deduced using the GP toolbox GPML's~\cite{Rasmussen2010a} conjugate gradients based minimization method for 1,000 iterations. 

\begin{algorithm}[tb]
	\caption{Surrogate-assisted Phenotypic Niching}
	\label{alg:SPHEN}
	\begin{algorithmic}
		\State Set $budget$, $maxGens$, $numInitSamples$ \Comment{Configure}
		\State $\mathcal{X}' \gets \mathcal{X} \gets Sobol(numInitSamples)$ \Comment{Initial samples}
		\While {$|\mathcal{X}| < budget$}
		    \State $(\mathbf{f}',\mathbf{p}') \gets Sim(\mathcal{X}')$ \Comment{Precisely evaluate performance and features}
		    \State $(\mathcal{X},\mathbf{f},\mathbf{p}) \gets (\mathcal{X} \cup \mathcal{X}', \mathbf{f} \cup \mathbf{f}',\mathbf{p} \cup \mathbf{p}')$ 
			\State $(\mathbf{M}_{p},\mathbf{M}_{f}) \gets Train(\mathcal{X},\mathbf{f},\mathbf{p})$ \Comment{Train surrogate models}
			\State $\mathbf{A}_{map} \gets \textsc{MAP-Elites}(20,\mathcal{X}, Predict(\mathcal{X},\mathbf{M}_{f}),Predict(\mathcal{X},\mathbf{M}_{p}),\mathbf{M}_{p},\mathbf{M}_{f})$ \Comment{Produce acquisition map based on predicted sample performance and features}
			\State $\mathcal{X}' \gets Sobol(\mathbf{A}_{map})$ \Comment{Select new (optimized) samples from acquisition map}

		\EndWhile
		\State $\mathbf{P}_{map} \gets \textsc{MAP-Elites}(acq(),0,feat(),\mathcal{X},\mathbf{f},\mathbf{p},\mathbf{M}_{p},\mathbf{M}_{f})$ \Comment{Produce prediction map}
		\State 
		\Procedure{MAP-Elites}{$\sigma_{ucb},\mathcal{X},\mathbf{f},\mathbf{p},\mathbf{M}_{p},\mathbf{M}_{f}$}		
		\State $\mathbf{I}_{map} \gets (\mathcal{X},\mathbf{f},\mathbf{p})$ \Comment{Create initial map}
		\While {$gens < maxGens$}
			\State $\mathbf{P} \gets Sobol(\mathbf{I}_{map})$ \Comment{Evenly, pseudo-randomly select parents from map}
			\State $\mathbf{C} \gets Perturb(\mathbf{P})$ \Comment{Perturb parents to get children}
			\State $\mathbf{f} \gets Predict(\mathbf{C},\mathbf{M}_{f})$ \Comment{Predict features}
			\State $\mathbf{p} \gets UCB(\mathbf{C},\sigma_{ucb},\mathbf{M}_{p})$ \Comment{Predict performance (Upper Confidence Bound)}
			\State $\mathbf{I}_{map} \gets Replace(\mathbf{I}_{map},\mathbf{C},\mathbf{f},\mathbf{p})$ \Comment{Replace bins if empty or better}
		\EndWhile
		\EndProcedure
	\end{algorithmic}
\end{algorithm}

MAP-Elites then creates the acquisition map by optimizing the UCB of the performance model (with a large exploration factor $\kappa = 20$), using feature models to assign niches for the samples and new solutions (Fig.~\ref{img:SPHEN}c). Notably, we do not take into account the confidence of those feature models. Surrogate assisted QD works, because, although the search takes place in a high-dimensional space, QD only has to find the \textit{elite hypervolume}~\cite{vassiliades2018}, or \textit{prototypes}~\cite{hagg2018prototype}, the small volumes consisting of high-performing solutions. Only the performance function can guide the search towards the hypervolume. Taking into account the feature models' confidence intervals adds unnecessary complexity to the modeling problem. SPHEN's goal is to be able to only predict features for high-performing solutions, so we let feature learning ``piggyback'' on this search. We use a Sobol sequence on the bins of the acquisition map to select new solutions (Fig.~\ref{img:SPHEN}d) that are then evaluated to continue training the surrogate models. This process iterates as long as the evaluation budget is not depleted. Finally, MAP-Elites is used to create a prediction map, ignoring the models' confidence (Fig.~\ref{img:SPHEN}e), which is filled with diverse, high-performing solutions.

\section{Domains}
\label{sec:domains}

Phenotypic features describe phenomena that can be related to complex domains, like behavioral robotics, mechanical systems, or computational fluid dynamics (CFD). Before we apply SPHEN to an expensive CFD domain, we compare its performance to MAP-Elites and SAIL in a simpler, inexpensive domain.

\subsection{Polygons}
\label{sec:domains:polygons}

\begin{figure}[tb]
	\centering
	\includegraphics{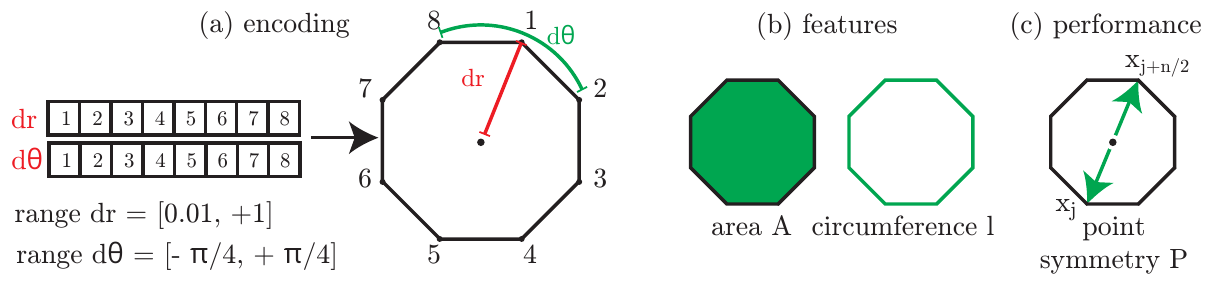}
	\caption{The genome (a), consisting of 16 parameters that define axial and radial deformations, shape features (b) and performance (c) of polygons in the domain.}
	\label{img:encoding}
\end{figure}

To be able to calculate all performance and feature values, we optimize free form deformed, eight-sided polygons. The polygons are encoded by 16 parameters controlling the polar coordinate deviation of the control points (Fig.~\ref{img:encoding}a). The first half of the genome determines the corner points' radial deviation ($dr \in [0.01,1]$). The second half of the genome determines their angular deviation ($d\theta \in [-\pi/4,\pi/4]$).
The phenotypic features are the area of the polygon $A$ and its circumference $l$ (Fig.~\ref{img:encoding}b). These values are normalized between $0$ and $1$ by using predetermined ranges ($A \in [0.01,0.6]$ and $l \in [1,4]$). The performance function (Fig.~\ref{img:encoding}c) is defined as the point symmetry $P$. The polygon is sampled at $n=100$ equidistant locations on the polygon's circumference, after which the symmetry metric is calculated (Eq.~\ref{eq}), based on the symmetry error $E_s$, the sum of Euclidean distances of all $n/2$ opposing sampling locations to the center:

\begin{equation}
f_P(x_i) = {1 \over {1 + E_s(x_i)} }, E_s(x) = \sum_{j=1}^{n/2}\left\lVert \vec{x}_j-\vec{x}_{j+n/2}\right\rVert 
\label{eq}
\end{equation}

\subsection{Air Flow}
\label{sec:domains:flow}

The air flow domain is inspired by the problem of wind nuisance in the built environment.
Wind nuisance is defined in building norms~\cite{janssen2013,NEN8100} and uses the wind amplification factor measured in standardized environments, with respect to the hourly mean wind speed. In a simplified 2D setup, we translate this problem to that of minimizing maximum air flow speed ($u_{Max}$) based on a fixed flow input speed. The performance is determined as the inverse over the normalized maximum velocity: $p(x) =  {2 \over (1 + u_{Max}(x))}-1$. However, we only need to keep $u_{Max}$ within a \textit{nuisance threshold}, which we set to $u_{Max} \leq 0.12$.

The encoding from the polygon domain is used to produce 2D shapes that are then placed into a CFD simulation. To put emphasis on the architectural nature of the domain, we use two features, area and air flow turbulence. The chaotic behavior of turbulence provokes oscillations around a mean flow velocity, which influences the maximum flow velocity. 
Both features are not optimization goals. Rather, we want to analyze, under the condition of keeping the flow velocity low, how the size of the area and turbulence are related to each other. We want to produce polygons that are combinations between their appearance (small to large) and their effect on the flow (low to high turbulence). Concretely, at the lowest and highest values of area and turbulence, regular intuitive shapes should be generated by the algorithm such as slim arrow-like shapes for low turbulence and area, or regular polygons for high turbulence and area. However, for area/turbulence combinations in between, the design of the shape is not unique and will possibly differ from intuition.

\subsubsection{Lattice Boltzmann Method}
\label{lbm}
The Lattice Boltzmann method (LBM) is an established tool for the simulation of CFD~\cite{Kruger2016}. Instead of directly solving the Navier-Stokes equations, the method operates a stream and collide algorithm of particle distributions derived from the Boltzmann equation. In this contribution, LBM is used on a 2D grid with the usual lattice of nine discrete particle velocities. At the inlets and outlets, the distribution function values are set to equilibrium according to the flow velocity. The full bounce-back boundary condition is used at the solid grid points corresponding to the polygon. Although there are more sophisticated approaches for the boundaries, this configuration is stable throughout all simulations. In addition, the bounce-back boundary condition is flexible, as the boundary algorithm is purely local with respect to the grid points.
As an extension of the Bhatnagar-Gross-Krook (BGK) collision model~\cite{Bhatnagar1954}, a Smagorinsky subgrid model~\cite{Gaedtke2018} is used to account for the under-resolved flow in the present configuration. For a more detailed description of the underlying mechanisms, we refer to~\cite{Kruger2016}. Note that the results of the 2D domain do not entirely coincide with results that will be found in 3D, caused by the difference in turbulent energy transport~\cite{Tennekes1978}. 

\begin{figure}[tb]
	\centering
	\includegraphics{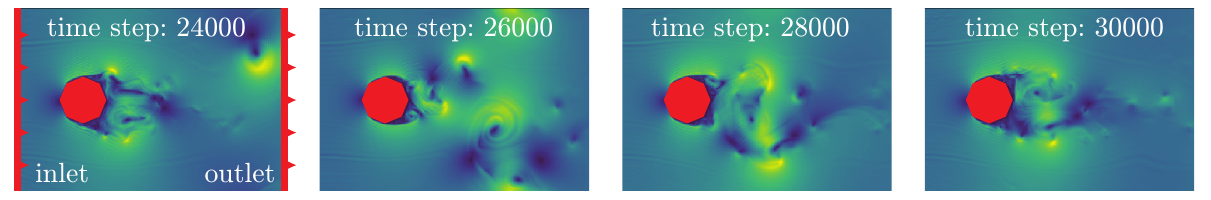}
	\caption{ Air flow around a circular polygon shape at four different time steps.}
	\label{img:flowexamples}
\end{figure}

The simulation domain consists of $300\cdot200$ grid points. A bitmap representation of the polygon is placed into this domain, occupying up to $64\cdot64$ grid points. As the Lattice Boltzmann method is a solver of weakly compressible flows, it is necessary to specify a Mach number (0.075), a compromise between computation time and accuracy. The Reynolds number is $Re=10,000$ with respect to the largest possible extent of the polygon. For the actual computation, the software package \emph{Lettuce} is used~\cite{Kramer2020}, which is based on the PyTorch framework~\cite{pytorch}, allowing easy access to GPU functionality. The fluid dynamics experiment was run on a cluster with four GPU nodes, each simulation taking ten minutes. Fig.~\ref{img:flowexamples} shows the air flow around a circular polygon at four different, consecutive time steps. Brighter colors represent higher magnitudes of air flow velocity. Throughout the 100,000 time steps of the simulation, maximum velocity and enstrophy are measured. The enstrophy, a measure for the turbulent energy dissipation in the system with respect to the resolved flow quantities~\cite{Gassner2013,Kramer2019}, increases as turbulence intensity increases in the regarded volume.

\subsubsection{Validation and Prediction of Flow Features}

\begin{figure}[tb]
	\centering
	\includegraphics{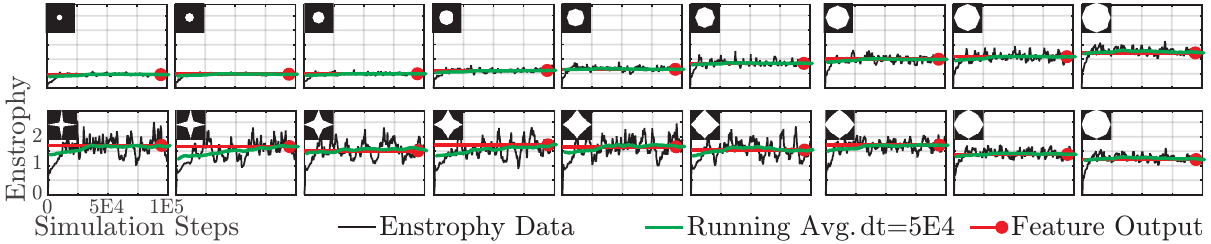}
	\caption{Enstrophy values during simulation of circles and stars. The running average of the last 50,000 time steps converges to the final feature output.}
	\label{img:convergence}
\end{figure}

The maximum velocity $u_{max}$ and enstrophy $E$ are measured every 50 steps. We employ a running average over the last 50,000 time steps. To test whether we indeed converge to a stable value, we run simulations with different shapes (nine varied-size circles and nine deformed star shapes) and calculate the moving average of the enstrophy values, which is plotted in Fig.~\ref{img:convergence}. The value converges to the final feature value (red). 

\begin{figure}[tb]
	\centering
	\includegraphics{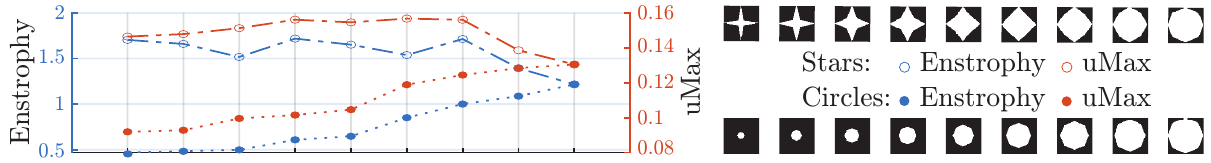}
	\caption{Enstrophy and maximum velocity of circles and stars.}	
	\label{img:validation}
\end{figure}

To validate the two measures, we simulate two small shape sets of circles and stars. Increasing the radius of the circles set should lead to higher $u_{max}$ and $E$, as more air is displaced by the larger shapes. The stars set is expected to have larger $u_{max}$ and $E$ for the more irregular shapes. This is confirmed in Fig.~\ref{img:validation}. 

Next, we investigate whether we can predict the simple shapes' flow feature values correctly. Although GP models are often called ``parameter free'', this is not entirely accurate. The initial guess for the hyperparameter's values, before minimization of the negative log likelihood of the model takes place, can have large effects on the accuracy of the model. The log likelihood landscape can contain local optima. We perform a grid search on the initial guesses for length scale and signal variance. Using leave-one-out cross validation, GP models are trained on all but one shape, after which we measure the accuracy using the mean absolute percentage error (MAPE), giving a good idea about the magnitude of the prediction error. The process is repeated until all examples were part of the test set once. The MAPE on $u_{Max}$ was $2.4\%$ for both sets. The enstrophy was harder to model, at $4.9\%$ and $10.3\%$ for the respective sets, but still giving us confidence that these two small hypervolumes can be modeled.

\section{Evaluation}
\label{sec:evaluation}

We evaluate how well SPHEN performs in comparison to SAIL and MAP-Elites when we include the cost of calculating the features, how accurate the feature models are when trained with a performance based acquisition function, and whether we can apply SPHEN to an expensive domain.

\subsection{Quality Diversity Comparison}

\begin{table}[bt]
	\centering
	\caption{Parameter settings for MAP-Elites, SAIL\textsuperscript{A}, restricted SAIL\textsuperscript{B} and SPHEN.}
	\label{tbl:parameter}
	{\setlength{\tabcolsep}{1em}
		\begin{threeparttable}
			\begin{tabular}{lrrrr}
				\textbf{Parameter} & MAP-Elites & SAIL\tnote{A}& SAIL\tnote{B} & SPHEN\\
				\hline
				Generations			& 4,096 	& 1,024 	& 63 	& 1,024	\\
				Descendants			& 16 	& 32 	& 16 	& 32	\\
				\hline
				Budget (per iteration)& - 	& 1,024 (16)	& 1,024 (16)	& 1,024 (16)	\\
				Resolution (acquisition)	& - 	& 16x16	& 16x16	& 32x32 \\
				\hline
			\end{tabular}
			\begin{tablenotes}\footnotesize
				\item[A] Due to the number of feature evaluations, MAP-Elites uses $4{,}096\cdot16=65{,}536$ and SAIL uses $16 + 1{,}024\cdot32\cdot({1{,}024 \over 16}) + 1{,}024 = 2{,}098{,}192$ evaluations.
				\item[B] Here, SAIL is restricted to the number of evaluations that was used in MAP-Elites. Number of generations (${4{,}096\cdot16-1{,}024-16 \over 1{,}024} \approx 63$).
			\end{tablenotes}
	\end{threeparttable}}
\end{table}

We run QD optimization without (MAP-Elites) and with surrogate model(s) (SAIL, SPHEN) on the polygon domain (Section~\ref{sec:domains:polygons}). This allows us to check all ground truth performance and feature values in a feasible amount of time. The shape features should be easier to learn than the flow features of the air flow domain. The working hypothesis is that we expect SPHEN to perform somewhere between SAIL and MAP-Elites, as it has the advantage of using a surrogate model but also has to learn two phenotypic features. But in the end, since the ultimate goal is to be able to use QD on expensive features, SPHEN will be our only choice. The parameterization of all algorithms is listed in Table~\ref{tbl:parameter}. The initial sample set of 16 examples as well as the selection of new samples (16 in every iteration) is handled by a pseudo-random Sobol sequence. The mutation operator adds a value drawn from a Gaussian distribution with $\sigma = 10\%$. 

\begin{figure}[tb]
	\centering
	\includegraphics{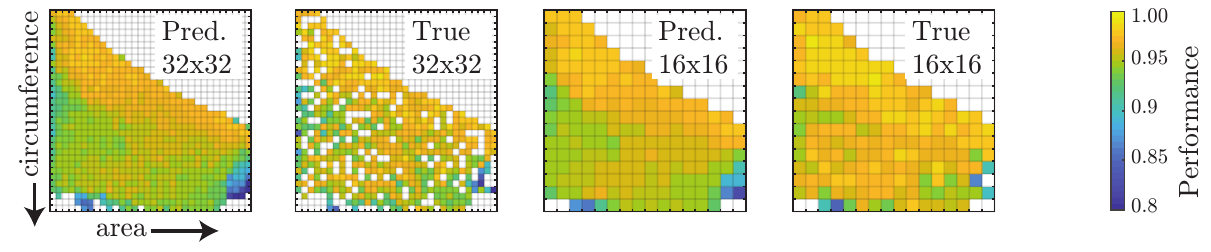}	
	\caption{Predicted and true SPHEN maps on symmetry domain, trained in 32x32 resolution (left), then reduced to 16x16 resolution  to remove holes (right).}
	\label{img:highlowres}
\end{figure}

Due to the expected inaccuracy of the feature models, misclassifications will decrease the accuracy of the maps. Fig.~\ref{img:highlowres} shows a prediction map at a resolution of 32x32 and the true performance and feature map. Holes appear due to misclassifications, which is why we train SPHEN on a higher resolution map and then reduce the prediction map to a resolution of 16x16. Most bins are now filled. In this experiment all prediction maps have a resolution of 16x16 solutions.

\begin{figure}[tb]
	\centering
	\includegraphics{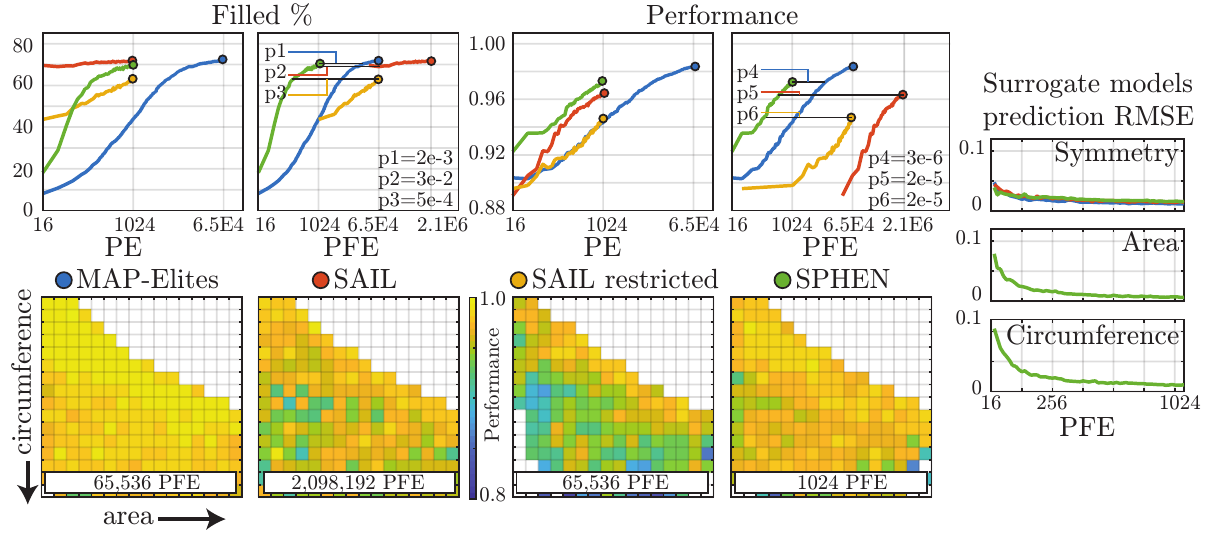}	
	\caption{Comparison of MAP-Elites, SAIL and SPHEN based on performance evaluations (PE) and performance/feature evaluations (PFE). Experiments were repeated five times to produce the mean percentage of map filled and mean performance values. Prediction errors are included on the right and example prediction maps at the bottom. The experiments include a version of SAIL that is restricted to the number of PFE used in MAP-Elites.}
	\label{img:qdcomparison}
\end{figure}

The mean amount of filled map bins and performance values for five replicates are shown in Fig.~\ref{img:qdcomparison}. SAIL and SPHEN find about the same number of solutions using the same number of performance evaluations (PE). Notably, the mean performance of SPHEN's solutions is higher than that of SAIL. However, in domains with expensive feature evaluations we need to take into account the performance or feature evaluations (PFE). SAIL now needs more than two million PFE to perform almost as well as SPHEN, which only needs 1,024, which is over three orders of magnitude less and still more than an order of magnitude less than MAP-Elites. Since in expensive real world optimization problems we cannot expect to run more than about 1,000 function evaluations, due to the infeasibly large computational investment, the efficiency gain of SPHEN is substantial. If we lower the number of PFE of SAIL to the same budget as MAP-Elites and give it more time to search the iteratively improving surrogate model before running out of the budget of 65,536 PFE (see Table~\ref{tbl:parameter}), SAIL still takes a big hit, not being able to balance out quality and diversity. The example prediction maps are labeled with the number of PFE necessary to achieve those maps. Although we do not sample new training examples to improve the feature models specifically, their root mean square error (RMSE) ended up at $0.012$ and $0.016$ respectively. Finally, we test SPHEN to the three alternative algorithms on the null hypothesis that they need the same number of PFE to reach an equally filled map or equal performance. Significance levels, calculated using a two-sample t-test, are shown in Fig.~\ref{img:qdcomparison}. In all cases, the null hypothesis is improbable ($p < 0.05$), although for the comparison of filled levels to SAIL it is rejected with less certainty.

We conclude that we do not need to adjust the acquisition function. SPHEN and SAIL search for the same elite hypervolume, which is only determined by the performance function. 

\subsection{Designing Air Flow}

After showing that SPHEN can learn both performance as well as feature models, we now run SPHEN in the air flow domain (Section~\ref{sec:domains:flow}). The objective is to find a diverse set of air flows using a behavioral feature, turbulence, and one shape feature, the surface area of the polygon. 
We want to find out how the size of the area and turbulence are related to each other and which shapes do not pass the wind nuisance threshold.
We use the same parameters for SPHEN as were listed in Table~\ref{tbl:parameter}, but allow 4,096 generations in the prediction phase. The enstrophy and velocity are normalized between $0$ and $1$ using a predetermined value range of $E \in [0.15,1.1]$ and $u_{Max} \in [0.05,0.20]$.

\begin{figure}[tb]
	\centering
	\includegraphics{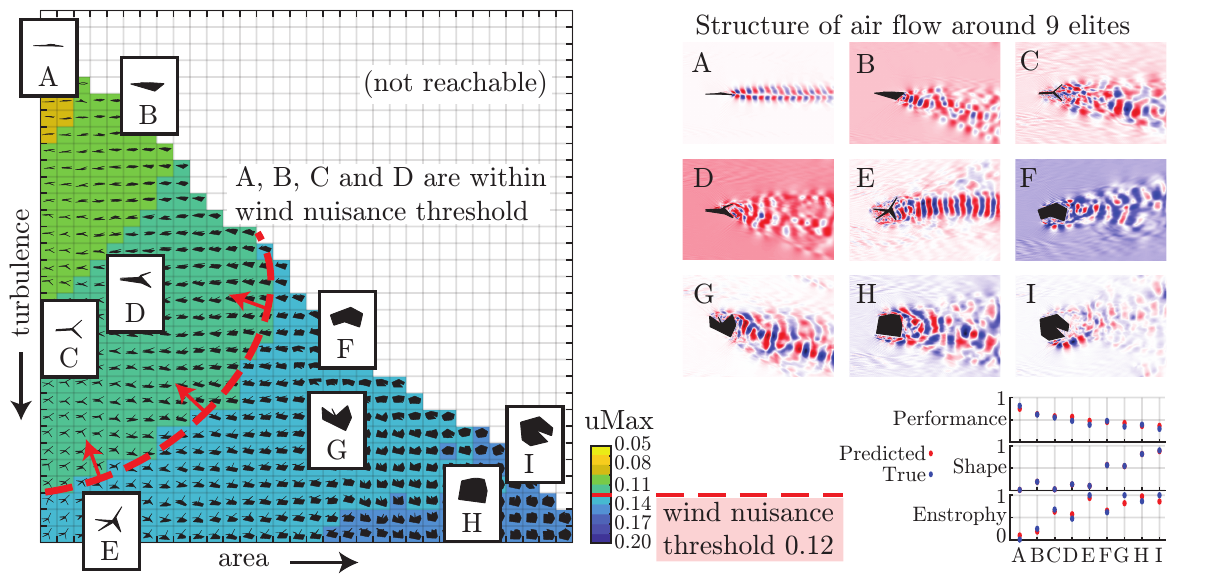}
	\caption{A diversity of shapes and air flows that shows which designs conform to the wind nuisance threshold. The dominant DMD mode shows the structure of the air flow around nine selected shapes. A, B, C and D are within the wind nuisance threshold.}
	\label{img:finalSimulations}
\end{figure}

The resulting map of solutions in Fig.~\ref{img:finalSimulations} shows that turbulence and surface area tend to increase the mean maximum air flow velocity, as expected. A small selection of air flows is shown in detail. RMSE of the models is 0.06, 0.01 and 0.10 respectively and Kendall's tau rank correlation to the ground truth amounts to 0.78, 1.00 and 0.73 (1.00 for A, B, C and D).

Due to the chaotic evolution of turbulent and transient flows, a static snapshot of the velocity field provides only limited information about the flow structures. Therefore, dynamic mode decomposition (DMD) is used to extract and visualize coherent structures and patterns over time from the flow field~\cite{PyDMD,Schmid2010}.

Especially those shapes at the extrema of area and turbulence align with the aerodynamic expectations as detailed in Section~\ref{sec:domains:flow}. At low turbulence intensity, the shapes tend to be slim and long with respect to the flow direction (shapes A and B). High turbulence levels at small shape areas are achieved if the shapes are oriented perpendicularly to the flow (shape E). Pentagons or hexagons evoke high turbulence levels at large areas (shapes H and I). However, impressively, there is an enormous variety of nuances in between these extrema with non-intuitive shapes, enabling the designer to determine a shape for given flow parameters down to a lower turbulence bound for each area value. Furthermore, the algorithm also suggests known tricks to manipulate the flow. Side arms are an appropriate measure to vary the turbulence intensity in the wake (shapes C, D, E, and G). Indentations or curved shapes redirect the flow and extract kinetic energy similar to turbine blades \cite{Dorschner2017}, which can be observed in shape D. Conclusively, for the highest and lowest area and turbulence values, SPHEN matches the expectations while for the shapes in between SPHEN exceeds expectations by introducing unusual shape nuances, which encourage further investigation.

\subsection{Discussion}

In the polygon domain, both surrogate-assisted algorithms are able to find a large variety of solutions.
When features do not have to be modeled, they show similar performance, although SAIL converges much sooner. 
However, when taking into account the number of feature evaluations, SPHEN clearly outperforms SAIL as well as MAP-Elites. 
Modeling features does not lower the performance of a prediction map. In terms of solution performance, both surrogate-assisted algorithms are outperformed by MAP-Elites in the simple domain, but SPHEN clearly beats MAP-Elites by requiring less evaluations. The feature models become more accurate even when sampling only to improve the performance model. 

When designing diverse air flows, one SPHEN run took 23 hours, producing 494 different air flow profiles. With SAIL, obtaining the same result would have taken over five years. Although MAP-Elites outperformed SAIL in the simple polygon domain, and might have outperformed it in the air flow domain as well, it still would have taken two months to calculate with uncertain result. Fig.~\ref{img:finalSimulations} shows that we can find structure in the air flows that can appear in this problem domain. We can efficiently combine variations (area) of the object we want to design as well as their effect on the environment (turbulence). Even when only using two phenotypic features, the nuances between the variations give us an idea which shapes do not pass the wind nuisance threshold and which ones do, and could continue the design process based on our new intuition.

\section{Conclusion}
\label{sec:conclusion}

In this work we showed that expensive phenotypic features can be learned along with an expensive performance function, allowing SPHEN, an evolutionary QD algorithm, to find a large diversity of air flows. In an inexpensive domain we showed that, when we take into account the number of feature evaluations, SPHEN clearly outperforms state of the art algorithms like MAP-Elites and SAIL. The result clears the way for QD to find diverse phenotypes as well as behaviors in engineering domains without the need for an infeasible number of expensive simulations. This is made possible because only the elite hypervolume needs to be modeled. Fluid dynamics domains count as some of the most complicated. Although often solved in ingenious ways by engineers relying on experience, QD can add \textit{automated intuition} to the design process. Variations of the object we want to optimize as well as variations in the effects on the object's environment can be seen ``at a glance'', which is what intuition is all about. 

The most urgent future work is to study whether we can make adjustments to the acquisition function, taking into account feature models' confidence intervals to improve SPHEN. Furthermore, the solution diversity should be analyzed in higher-dimensional feature spaces and applied to 3D shapes.
\\ \\
We showed what expected and unexpected behavioral patterns can emerge in complicated problem domains using surrogate-assisted phenotypic niching. Our main contribution, automatic discovery of a broad intuition of the interaction between shape and behavior, allows engineers to think more out-of-the-box.

\subsubsection*{Acknowledgments.} 
This work was funded by the Ministry for Culture and Science of the state of Northrhine-Westphalia (grant agreement no. 13FH156IN6) and the German Research Foundation (DFG) project FO 674/17-1. The authors thank Andreas Kr\"amer for the discussions about the Lettuce solver. 

\bibliographystyle{splncs04}
\bibliography{qd,lbm}
\end{document}